\documentclass[10pt,twocolumn,letterpaper]{article}

\usepackage[pagenumbers]{wacv} % To force page numbers, e.g. for an arXiv version

\usepackage{booktabs}
\usepackage{multirow}
\usepackage{array}

\newcommand{\imgwidth}{0.198\linewidth}

\usepackage{tikz}
\usetikzlibrary{positioning, arrows.meta, calc, fit, backgrounds}

\definecolor{frozenfill}{RGB}{226,232,240}  % frozen (grey-blue)
\definecolor{trainfill}{RGB}{255,237,213}   % trainable (amber)
\definecolor{hfamfill}{RGB}{220,242,224}     % HFAM sub-blocks (green)
\definecolor{gtgrey}{RGB}{120,120,120}       % training-only path

\tikzset{
  fbox/.style={rounded corners=2pt, draw, semithick, align=center,
               font=\footnotesize, inner sep=3pt, minimum height=7mm},
  frozen/.style={fbox, fill=frozenfill},
  train/.style={fbox, fill=trainfill},
  hfam/.style={fbox, fill=hfamfill},
  lbl/.style={font=\footnotesize, inner sep=1pt},
  flow/.style={-{Latex[length=1.8mm]}, semithick},
  flowb/.style={{Latex[length=1.8mm]}-{Latex[length=1.8mm]}, semithick},
  gtflow/.style={-{Latex[length=1.8mm]}, semithick, dashed, draw=gtgrey},
  losstxt/.style={font=\scriptsize, align=left, text=gtgrey},
}

\definecolor{wacvblue}{rgb}{0.21,0.49,0.74}
\usepackage[pagebackref,breaklinks,colorlinks,allcolors=wacvblue]{hyperref}
\usepackage[accsupp]{axessibility}  % Improves PDF readability for those with disabilities.

\def\wacvPaperID{644} % *** Enter the WACV Paper ID here
\def\confName{WACV}
\def\confYear{2027}

\title{CLEAR: Color-Texture Latent Alignment for \\ Single Image Reflection Removal}

\author{
Tu Vo \quad Sheir A. Zaheer \quad Chan Y. Park\\
KC Machine Learning Lab\\
{\tt\small tuvv@kc-ml2.com, sheir@kc-ml2.com, chan.y.park@kc-ml2.com}
}

\begin{document}
\maketitle
\begin{abstract}
Single Image Reflection Removal (SIRR) seeks to recover the transmitted scene from a photograph captured through glass. Recent latent restoration methods use pretrained generative priors, but many recover detail by feeding high-frequency features from the degraded input back into the decoder. This works for degradations such as blur, noise, or haze, where input detail mostly belongs to the scene. It is less suitable for SIRR, where reflections add a second structured layer; directly reusing the input's skip connections can sharpen the image while also bringing reflection edges back. Latent compression also introduces color drift and fine-detail loss. We propose \textbf{CLEAR} (\textbf{C}olor-Texture \textbf{L}at\textbf{E}nt \textbf{A}lignment for Single Image \textbf{R}eflection Removal), a deterministic, single-timestep, non-iterative latent framework that uses a pretrained latent diffusion backbone as a natural-image prior. A physics-guided separator organizes the prediction around the \emph{transmission, reflection, and residue} formation model and outputs coarse layer latents in one forward pass. The \textbf{Hybrid Feature Alignment Module (HFAM)} then treats input-derived skips as contaminated layer-mixture features. It rectifies their high-frequency content with a clean-transmission-aligned color query, learned from the input features themselves under ground-truth feature supervision, keeping transmission-consistent texture while attenuating reflection-colored detail. CLEAR achieves the best average across fidelity and perceptual metrics on six real-world benchmarks and the lowest LPIPS on every dataset, with the largest gains on the hardest benchmarks.
\end{abstract}

\vspace{-1mm}
\section{Introduction}
\label{sec:intro}
Photographs taken through glass often contain unwanted reflections, where a mirrored scene is overlaid on the intended background. The captured image is the sum of a transmission layer (the background) and a reflection layer (the mirrored scene). Single Image Reflection Removal (SIRR) aims to recover the clean transmission from this mixture~\cite{dong2021location, fan2017generic, hong_ijcv_2024, li2020single, wan2018crrn, wei2019single, zhang_2022, hong2024ldiffer, hu2025dereflection}. Clean transmission improves everyday photography and also helps downstream vision tasks such as detection, segmentation, and recognition, whose accuracy drops when the input carries reflections. The task is ill-posed: many transmission--reflection pairs can produce the same observation~\cite{hong2024ldiffer}, so any method must rely on learned priors.

CNN-based methods~\cite{hu2023single, zhu2024revisiting, song2023robust, hong_ijcv_2024, hu2024single, zhang2018single} work well with large datasets and semantic priors, but tend to over-smooth heavily contaminated regions and lose high-frequency detail. To recover this detail, recent work builds on pre-trained latent generative models~\cite{ho2020denoising, song2021denoising, hong2024ldiffer, rombach2022high}, which carry natural-image priors and can fill in plausible texture where the input is ambiguous.

These generative approaches treat reflection removal as a generic conditional restoration problem: they condition on mixed input and recover detail by reusing input-derived features during decoding~\cite{hu2025dereflection, hong2024ldiffer, windowseat2025}. Latent restoration decoders make this explicit. SLURPP~\cite{wu2025single}, for example, restores high-frequency detail by adding skip connections from the degraded input image into the decoder (\emph{input skips}). This is the right choice when the degradation removes information: in underwater imaging, blur, or noise, the input's high-frequency content mostly belongs to the scene, so copying it back recovers real detail.

Reflection removal is the case where this fails. Reflection does not remove information; it \emph{adds} extraneous structure, and the input's high-frequency content includes the reflection's own edges. Feeding the input back into the decoder therefore reintroduces the reflections that the network just removed, producing residual ghosting and structural inconsistencies~\cite{fan2017generic, wan2018crrn, yang2018seeing, hu2023single}. Decoder-side detail recovery for SIRR is therefore more than a sharpening problem. The skip stream is itself a layer mixture, and the model must decide which high-frequency channels belong to the transmission before they enter the decoder. Working in a compressed latent space adds two additional problems: the high VAE compression ratio shifts colors, and the regression objective loses fine detail.

CLEAR addresses these issues with a deterministic, single-timestep, non-iterative latent framework that uses a pretrained latent diffusion backbone as a natural-image prior (no random sampling, no noise schedule) and pairs it with explicit transmission--reflection decomposition. CLEAR is a two-stage pipeline: a dual-branch latent separator (Stage~1) and a decoder augmented with the Hybrid Feature Alignment Module (HFAM, Stage~2). Stage~1 is a physics-guided latent separator for single-image SIRR that predicts transmission, reflection, and residue latents in one forward pass under the residue-based image formation of DSRNet~\cite{hu2023single}; its two branches exchange information through latent cross-attention to suppress correlated, entangled features. Following DSRNet, the residual is tied to transmission and reflection only through the reconstruction loss, not through an explicit function.

This coarse transmission is largely reflection-free but, as noted above, carries a color shift from latent compression and has lost fine detail. The decoder restores that detail from the input's skip features, but those features carry the reflection. At each decoder scale, HFAM learns to transform the input-derived high-frequency skips toward clean-transmission feature statistics, using the ground-truth transmission features only as training targets. The transform is driven by a color query that HFAM computes from the input skips but trains to match the clean transmission's color, so transmission-consistent texture is kept, reflection-colored detail is attenuated, and the latent color shift is corrected. Because the detail is aligned to the clean target rather than copied from the input, the reflection is suppressed rather than reintroduced.

Our contributions are:
\begin{itemize}
    \item We instantiate the explicit \emph{transmission--reflection--residue} formation model ($\mathbf{I}=\mathbf{T}+\mathbf{R}+\Phi$) as the \emph{prediction target} of a deterministic latent separator, jointly predicting the transmission, reflection, and a learned residue ($F_T, F_R, F_\Phi$) in a single forward pass for single-image reflection removal.
    \item We show that direct input-skip reuse, effective for information-loss degradations, reintroduces reflections in SIRR because the input's high-frequency content includes the reflection's own edges. HFAM aligns contaminated input skips toward clean-transmission feature statistics before they enter the decoder.
    \item We test the mechanism on six real-world benchmarks, sub-path ablations, global color-fix comparisons, parameter-matched backbones, and a full-reference reflection-leakage metric with paired-bootstrap confidence intervals. The full method improves PSNR by 0.61\,dB, SSIM by 0.037, and LPIPS by 33\% (0.159$\to$0.106) over the strongest single-timestep baseline, DAI.
\end{itemize}

\vspace{-1mm}
\section{Related Work}
\label{sec:related}

\subsection{Single Image Reflection Removal}
SIRR methods fall broadly into optimization-based and deep learning-based approaches.

\noindent\textbf{Optimization-based methods.}
Early works relied on handcrafted priors to distinguish the transmission layer from the reflection. Levin and Weiss~\cite{levin2007user} separated reflections using gradient sparsity priors. Li and Brown~\cite{li2014single} used relative smoothness, assuming that the reflection is more out-of-focus than the transmission. Shih~\etal~\cite{shih2015reflection} exploited ghosting cues from double-pane glass, and others explored convex optimization~\cite{yang2019fast} and region-aware priors~\cite{wan2018region}. While foundational, these methods struggle with sharp reflections and complex non-linear superpositions~\cite{wen2019single}.

\noindent\textbf{Deep learning-based methods.}
Following the first generic deep architecture for layer separation~\cite{fan2017generic}, many works added reflection-specific modules: multi-scale and cooperative edge guidance (CRRN~\cite{wan2018crrn}, CORRN~\cite{wan2019corrn}), misaligned-data training (ERRNet~\cite{wei2019single}), cascaded refinement (IBCLN~\cite{li2020single}), perceptual losses~\cite{zhang2018single}, and location-aware processing~\cite{dong2021location}. The dual-stream paradigm of YTMT~\cite{hu2021trash} and DSRNet~\cite{hu2023single} is most relevant to us; DSRNet models a learnable residue to capture non-linearities in the superposition, and we follow this residue formulation. Recent methods extend robustness and capacity: adversarial robustness (RobustSIRR~\cite{song2023robust}), the large-scale RRW dataset~\cite{zhu2024revisiting}, transformer attention (DSIT~\cite{hu2024single}), reversible layers (RDNet~\cite{zhao2025reversible}), and inter-layer complementarity (ILCT)~\cite{huang2025single}, alongside work on data synthesis and physical modeling~\cite{kim2020single, zheng2021single, zheng2020what}. More recently, Kee~\etal~\cite{kee2025removing} remove reflections end-to-end from RAW photographs, optionally using an opposite-facing context photo; complementary multi-cue setups (flash/no-flash~\cite{flashsplit}, polarization) ease separation but require specialized captures, whereas CLEAR takes a single display-referred RGB image. Our dual-branch decomposition shares this stream-based lineage but operates in a compressed generative latent space.

\subsection{Generative Priors for Image Restoration}
Pre-trained generative models~\cite{rombach2022high} are now useful restoration priors. Pixel-space conditional methods (SR3~\cite{saharia2022image}, Palette~\cite{saharia2022palette}) concatenate the degraded input with the latent to guide an iterative reverse process. In reflection removal, L-DiffER~\cite{hong2024ldiffer} guides iterative denoising with text descriptions that lack spatial precision, while Flash-Split~\cite{flashsplit} uses iterative dual-branch latent diffusion with paired flash/no-flash captures.

A separate line of work maps the input to the output in one step~\cite{song2023consistency, sauer2024adversarial, yue2024resshift, wang2025osdface}. For reflection removal, DAI~\cite{hu2025dereflection} uses a single-timestep backbone conditioned on mixed input via ControlNet~\cite{zhang2023adding}, which leaves residual reflections in heavily corrupted regions; WindowSeat~\cite{windowseat2025} adapts a large DiT~\cite{dit} but its stochastic objective tends to oversmooth content. Deterministic single-timestep latent decomposition has also been used for underwater restoration~\cite{wu2025single}. CLEAR instead makes the explicit transmission/reflection/residue decomposition the network's prediction target, and its decoder changes the role of input skips: instead of copying them as high-frequency detail, HFAM aligns their detail to the clean transmission (\cref{sec:method}).

\noindent\textbf{Latent-space alignment and decoder refinement.}
A known failure mode of VAE-based latent restoration~\cite{kingma2013auto, rombach2022high} is that the decoder shifts colors and loses high-frequency detail. Existing remedies are mostly global or post-hoc. StableSR~\cite{colorfix} applies wavelet- and AdaIN-based color correction after decoding, following classic appearance transfer that matches channel-wise feature statistics~\cite{adain, gatys2016, reinhard2001}. Learned feature modulation, such as FiLM~\cite{film}, conditions features with per-channel scale and shift parameters. These operate on global statistics or apply a single transform across space, so they can fix overall color but cannot decide \emph{which} spatial textures to keep. HFAM instead performs a learned color-texture alignment inside the decoder: it learns a clean-transmission-aligned chromatic signature from the input-derived skips and uses it to reweight high-frequency subbands~\cite{dwt} channel-wise at each scale, so color correction and texture selection are coupled rather than handled separately. We compare against these global color-fix baselines in \cref{sec:exp}. The residue term in our layer model is a learned residual in the spirit of DSRNet~\cite{hu2023single}, tied to the layers only through the reconstruction loss. Two choices separate our approach from prior latent-prior SIRR: we align the input-derived skip detail to the clean transmission rather than copying it in, and HFAM targets the chromatic and high-frequency degradations specific to latent regression.

\vspace{-1mm}
\section{Method}
\label{sec:method}

\subsection{Preliminaries}

We distinguish a \emph{Stage} (a sequential pipeline component: Stage 1 disentangles the latent mixture, Stage 2 refines it via HFAM) from a \emph{Step} (the number of forward evaluations $N$); CLEAR runs $N{=}1$ without a sampling loop (details in the supplement).

\noindent\textbf{Image formation model.}
The linear model $\mathbf{I} = \mathbf{T} + \mathbf{R}$~\cite{fan2017generic} is simple but ignores light attenuation through glass~\cite{SIR2}. Weighted~\cite{wan2018crrn, yang2018seeing} and spatial alpha-matting variants ($\mathbf{I} = \mathbf{W}\circ\mathbf{T} + (1{-}\mathbf{W})\circ\mathbf{R}$) improve realism but require a hard-to-estimate pixel-wise map $\mathbf{W}$. Following DSRNet~\cite{hu2023single}, where a residual non-linearity is shown to be easier to estimate than a per-pixel $\mathbf{W}$ and to outperform the linear reconstruction, we adopt the residue-based formation:
\begin{equation}
\mathbf{I} = \mathbf{T} + \mathbf{R} + \Phi,
\label{eq:formation}
\end{equation}
where $\Phi$ is a learned residual term that absorbs non-linear deviations from the additive base. Following DSRNet, $\Phi$ is \emph{predicted} by the network, not computed as an explicit function of $\mathbf{T}$ and $\mathbf{R}$; it is tied to them only through the reconstruction constraint in \cref{eq:loss_1}; the additive layers then capture the dominant content and the residual absorbs non-linear effects such as saturation~\cite{hu2023single}. An audit over all 1{,}040 benchmark images shows $\Phi$ carries only 0.00082\% of the $R+\Phi$ L1 energy, is non-zero in 80\% of images, and is localized to saturated pixels (details in the supplement). The supplement shows an example decomposition into the recovered layers and residual.

\noindent\textbf{Latent generative priors.}
Stable Diffusion~\cite{rombach2022high} provides a pre-trained VAE latent space and U-Net prior. We reuse these weights only as a natural-image prior and train a direct deterministic restoration mapping; there is no noise schedule, no added Gaussian noise, and no iterative reverse chain. The supplement gives the exact single-timestep/SDv2 adaptation.

\subsection{Pipeline Overview}
\begin{figure*}[tb!]
    \centering
    \includegraphics[width=0.90\linewidth]{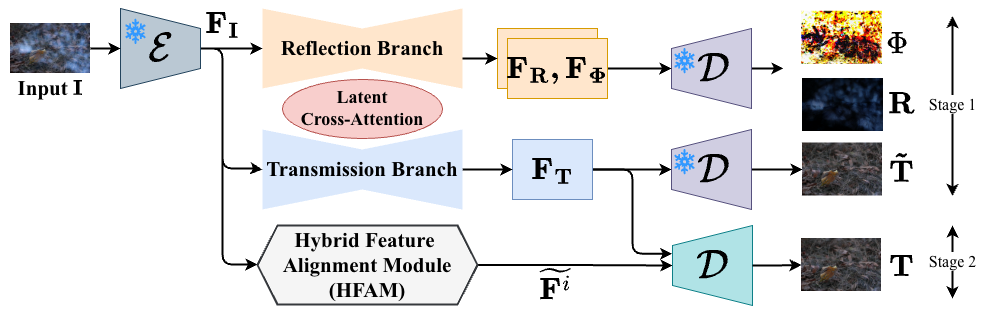}
    \caption{Pipeline overview. \textbf{Stage 1 (latent disentanglement):} the input is encoded by a frozen SDv2 VAE~\cite{rombach2022high} and processed by two U-Net branches that exchange information via latent cross-attention, predicting transmission and reflection/residue latents in a single deterministic step. \textbf{Stage 2 (feature alignment):} the transmission latent is decoded by a cross-latent decoder augmented at every scale by the HFAM, which injects high-frequency detail and chromatic cues. During training the HFAM is supervised with ground-truth features extracted by the same frozen encoder. Snowflakes mark frozen modules; trainable modules are unmarked.}
    \label{fig:pipeline}
\end{figure*}

As shown in \cref{fig:pipeline}, the contaminated image $\mathbf{I}$ is projected to a latent $F_I = \mathcal{E}(\mathbf{I})$ by the frozen VAE encoder. This latent feeds a \textbf{dual-path latent separator} of two symmetric U-Net branches initialized from identical pre-trained weights~\cite{rombach2022high}. Two-branch layouts are common in layer separation~\cite{hu2021trash, hu2023single}; related latent work has also used dual branches for paired flash/no-flash separation~\cite{flashsplit}. CLEAR uses this layout for a SIRR-specific latent prediction structure: one branch predicts the transmission latent $F_T$, while the other predicts the reflection and residue latents $F_R, F_\Phi$ so that the output follows the residue formation model in \cref{eq:formation}. The branches interact through latent cross-attention to identify and suppress correlated, entangled features. All latents are produced in one feed-forward pass through the dual-branch separator without iterative denoising steps or noise schedule.

\subsection{Stage 1: Latent Signal Disentanglement}
Stage 1 directly optimizes the dual-path network to decompose the latent mixture. Unlike noise-prediction models~\cite{ho2020denoising, hong2024ldiffer}, it performs a single-timestep latent projection of $F_I$ into $F_T, F_R, F_\Phi$ satisfying the residue constraint of \cref{eq:formation}. The objective combines per-layer fidelity, a self-supervised reconstruction term, and a gradient-domain exclusion term:
\begin{equation}
\label{eq:loss_1}
\mathcal{L}_{\text{S1}} = \lambda_T \mathcal{L}_{T} + \lambda_R \mathcal{L}_{R} + \lambda_{rec}\,\mathcal{L}_{rec} + \lambda_{exc}\,\mathcal{L}_{exc},
\end{equation}
where $\mathcal{L}_{rec} = \|\mathbf{I} - (\mathbf{T} + \mathbf{R} + \Phi)\|_1$ forces the predicted layers and residual to recombine into the input, which prevents plausible but unfaithful content. Each $\mathcal{L}_{k} = \mathcal{L}_{k}^{1} + \mathcal{L}_{k}^{\text{SSIM}} + \mathcal{L}_{k}^{\text{LPIPS}}$, $k\in\{\mathbf{T},\mathbf{R}\}$, supervises individual layer fidelity in pixel, structural, and perceptual terms. The exclusion loss $\mathcal{L}_{exc}$~\cite{zhang2018single} decorrelates transmission and reflection gradients to enforce pixel-level independence. For real data lacking ground-truth $\mathbf{R}$, the reflection terms are masked out.

\subsection{Stage 2: Decoder Refinement via HFAM}
\begin{figure}[tb!]
    \centering
    \includegraphics[width=\linewidth]{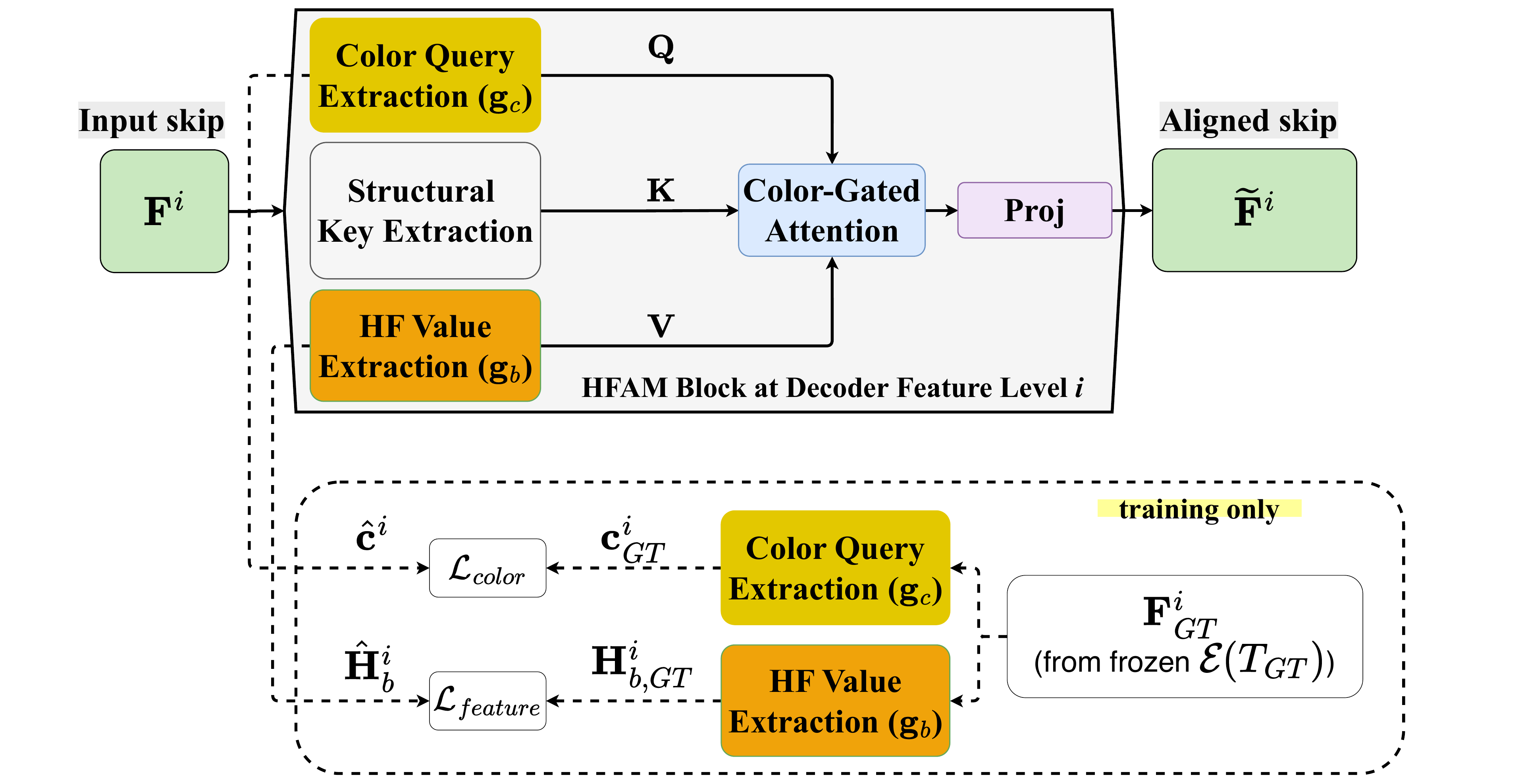}
    % \caption{HFAM at one decoder scale. From the input-derived skip feature $\mathbf{F}^{i}$ it forms a color query $Q$, structural key $K$, and high-frequency (DWT) value $V$; color-gated channel attention reweights $V$, keeping transmission-consistent texture and attenuating reflection-colored detail. \textbf{At inference, $Q$, $K$, $V$ are all computed from $\mathbf{F}^{i}$}; during training only (dashed), the predicted descriptors $\hat{\mathbf{c}}^{i}, \hat{\mathbf{H}}^{i}$ are supervised by frozen-encoder features $F_{GT}^{i}$ from the clean transmission, so the query learns the clean-scene color. HFAM is applied at every decoder scale.}
    \caption{HFAM at one decoder scale. Input skip $\mathbf{F}^i$ yields color query $Q$, structural key $K$, and high-frequency value $V$ (DWT); color-gated channel attention reweights $V$ toward clean-transmission statistics (supervised at training, computed from $\mathbf{F}^i$ at inference). Applied at all decoder scales.}
    \label{fig:hfam}
\end{figure}

\noindent\textbf{Motivation.}
Stage 1 yields a largely reflection-free transmission, but it is blurry: the high latent compression ratio and the deterministic objective both cost high-frequency detail, and decoding also drifts in color. Restoring that detail requires a high-frequency source, and the only adequate signal is the input image. This poses a dilemma specific to reflection removal. Direct input skips work when the degradation only removes information: the input's high-frequency content then belongs to the true scene and is merely attenuated, so copying it back restores real detail. For reflection, this no longer holds because the input's high-frequency content also contains the reflection's own edges, and copying it back reintroduces the reflection. HFAM turns skip injection into an alignment problem. It keeps the input skip-connections, whose detail is needed, but does not copy them directly. Instead, it \emph{learns to align} the input-derived skip features to the clean (ground-truth) transmission, reweighting high-frequency texture with a clean-transmission-aligned color query so that reflection-colored detail is attenuated before it reaches the decoder.

\noindent\textbf{Color-gated channel attention.}
HFAM is applied at each decoder scale $i\in\{1,\dots,4\}$, while the decoder is seeded by the Stage-1 transmission latent. From the input-derived skip feature $F^{i}\in\mathbb{R}^{C\times H\times W}$ it forms three role-specialized projections:
\begin{itemize}
    \item a \textbf{color query} $Q$ from a learned color descriptor $\hat{\mathbf{c}}^{i}$ of $F^{i}$ (a global channel-statistics chromatic signature; justified below), supervised during training to match the ground-truth transmission's color so the query carries the clean-scene color;
    \item a \textbf{structural key} $K$ from $F^{i}$ via $1{\times}1$ convolution;
    \item a \textbf{high-frequency value} $V$ from a single-level Haar discrete wavelet transform (DWT)~\cite{dwt} of $F^{i}$: the high-pass subbands $\{LH,HL,HH\}$ pass through a $1{\times}1$ convolution and are recombined by the inverse DWT to isolate edges and texture. The DWT/IDWT are fixed and differentiable; only the value, attention, and output projections are learned.
\end{itemize}
After $\ell_2$-normalizing $Q$ and $K$, HFAM applies a multi-head channel-wise (transposed) attention~\cite{restormer}:
\begin{equation}
\label{eq:hfam}
    \widetilde{F}^{i} = \mathrm{Proj}\!\left( \mathrm{Softmax}\!\big(\hat{Q}\,\hat{K}^{\top}\big)\, V \right), \qquad \hat{Q}\,\hat{K}^{\top}\in\mathbb{R}^{C\times C}.
\end{equation}
The affinity is computed \emph{across channels}, not across space (cost is linear in the number of pixels). Because the query is trained to the clean transmission color, the attention amplifies high-frequency texture favored by the clean-transmission-aligned query and attenuates the reflection-colored channels. A zero-initialized convolution then passes $\widetilde{F}^{i}$ into the decoder skip at scale $i$.

\noindent\textbf{Color descriptor.}
This color descriptor pools global channel statistics of $F^{i}$, following classic appearance-transfer results: spatial-mean channel statistics describe an image's global appearance with color as the dominant component (AdaIN~\cite{adain}; Gatys~\etal~\cite{gatys2016}), and per-channel global statistics suffice for color transfer (Reinhard~\etal~\cite{reinhard2001}). Because SDv2 VAE is trained to reconstruct RGB, its latent channel statistics correlate strongly with color, so the pooled descriptor is a compact chromatic signature.

\noindent\textbf{Alignment to the clean manifold.}
HFAM is trained to map these input-derived features to the clean transmission. We pass the ground-truth transmission through the same \emph{frozen} encoder $\mathcal{E}$ to obtain target features $F_{GT}^{i}$, used as supervision targets only. From the input skip $F^{i}$, HFAM \emph{predicts} an aligned color descriptor $\hat{\mathbf{c}}^{i}=g_c(F^{i})$ and high-frequency subbands $\hat{\mathbf{H}}_{b}^{i}=g_b(F^{i})$ with learned heads $g_c, g_b$. Applying the same heads to the GT feature gives the targets $\mathbf{c}^{i}_{\mathrm{GT}}=g_c(F_{GT}^{i})$ and $\mathbf{H}^{i}_{b,\mathrm{GT}}=g_b(F_{GT}^{i})$, against which we supervise two alignment losses at each scale. The GT-side descriptor outputs are detached (stop-gradient); the alignment losses update the shared heads $g_c,g_b$ through the input arm only.
\begin{align}
    \mathcal{L}_{color} &= \sum_{i} \big\| \hat{\mathbf{c}}^{i} - \mathbf{c}_{GT}^{i} \big\|_{1}, \\
    \mathcal{L}_{feature} &= \sum_{i}\sum_{b\in\{LH,HL,HH\}} \big\| \hat{\mathbf{H}}_{b}^{i} - \mathbf{H}^{i}_{b,\mathrm{GT}} \big\|_{1},
\end{align}
with $\mathcal{L}_{HFAM} = \mathcal{L}_{color} + \mathcal{L}_{feature}$. The targets $F_{GT}^{i}$ enter only these losses, never the decoded output; at inference, HFAM applies the learned $g_c, g_b$ to $F^{i}$ alone ($Q$, $K$, $V$ all derive from $F^{i}$). The full Stage 2 objective adds the decoded-transmission fidelity term $\mathcal{L}_{T}$ ($\ell_1$, SSIM, LPIPS):
\begin{equation}
\label{eq:loss_2}
  \mathcal{L}_{\text{S2}} = \lambda_{1}\mathcal{L}_{T} + \lambda_{2}\mathcal{L}_{HFAM}.
\end{equation}

\noindent\textbf{Relation to prior latent refinement.}
HFAM differs from existing decoder-side techniques in both what it aligns and how. Direct input-skip decoders copy input detail into the decoder unchanged; HFAM treats those skips as mixed-layer evidence and aligns their usable detail to the clean target manifold, so reflection edges are suppressed rather than reintroduced. Post-hoc color fixes~\cite{colorfix} and feature-modulation layers~\cite{film} match only global statistics or apply a single transform across space; HFAM's color-gated attention couples color correction to channel-wise texture selection. The color and high-frequency paths are also complementary, not redundant, as the ablation in \cref{sec:exp} confirms.

\subsection{Implementation Details}

% \begin{figure}[!b]
% \centering
% \setlength{\tabcolsep}{2pt}
% \renewcommand{\arraystretch}{0.6}
% \begin{tabular}{ccccc}
% \scriptsize Input & \scriptsize DSRNet & \scriptsize DSIT & \scriptsize RDNet & \scriptsize Ours \\
% \includegraphics[width=0.195\linewidth]{sup_R/input/test_0047_composite_img.png} &
% \includegraphics[width=0.195\linewidth]{sup_R/DSRNet/test_0047_R.png} &
% \includegraphics[width=0.195\linewidth]{sup_R/DSIT/test_0047_R.png} &
% \includegraphics[width=0.195\linewidth]{sup_R/RDNet/test_0047_R.png} &
% \includegraphics[width=0.195\linewidth]{sup_R/Ours/test_0047_r.png} \\
% \includegraphics[width=0.195\linewidth]{sup_R/input/test_0068_composite_img.png} &
% \includegraphics[width=0.195\linewidth]{sup_R/DSRNet/test_0068_R.png} &
% \includegraphics[width=0.195\linewidth]{sup_R/DSIT/test_0068_R.png} &
% \includegraphics[width=0.195\linewidth]{sup_R/RDNet/test_0068_R.png} &
% \includegraphics[width=0.195\linewidth]{sup_R/Ours/test_0068_r.png} \\
% \end{tabular}
% \caption{Predicted reflection layers ($R$). Compared with prior dual-stream methods, CLEAR preserves structure while reducing leakage of transmission content into the reflection estimate.}
% \label{fig:reflection_main}
% \end{figure}

\begin{table*}[t]
\centering
\caption{Quantitative comparison on six real-world benchmarks. $\uparrow$/$\downarrow$ indicate higher/lower is better. \textbf{Best} in bold, \underline{second-best} underlined.}
\label{tab:quantitative_comparison}
\small
\setlength{\tabcolsep}{4pt}
\renewcommand{\arraystretch}{1.05}
\begin{tabular}{cccccccccc}
\toprule
\multirow{2}{*}{\begin{tabular}{c}Dataset\\(size)\end{tabular}} & \multirow{2}{*}{Metric} & \multicolumn{8}{c}{Method} \\
\cmidrule(lr){3-10}
& & RobustSIRR & RRW & DSIT & DSRNet & RDNet & DAI & WindowSeat & Ours \\
\midrule
\multirow{5}{*}{\begin{tabular}{c}Nature\\(20)\end{tabular}}
& PSNR$\uparrow$ & 20.84 & 25.64 & 26.24 & 21.06 & 25.91 & \textbf{27.05} & \underline{27.00} & 26.45 \\
& SSIM$\uparrow$ & 0.766 & 0.832 & 0.831 & 0.799 & 0.832 & \underline{0.846} & 0.839 & \textbf{0.867} \\
& LPIPS$\downarrow$ & 0.229 & 0.178 & 0.190 & 0.194 & \underline{0.173} & 0.212 & 0.208 & \textbf{0.126} \\
& MANIQA$\uparrow$ & 0.481 & 0.457 & 0.461 & \underline{0.483} & 0.478 & 0.446 & 0.464 & \textbf{0.516} \\
& QUALICLIP$\uparrow$ & 0.720 & 0.708 & 0.715 & \underline{0.716} & 0.707 & 0.663 & 0.657 & \textbf{0.721} \\
\midrule
\multirow{5}{*}{\begin{tabular}{c}SIR$^2$\\(500)\end{tabular}}
& PSNR$\uparrow$ & 22.10 & 25.19 & 25.48 & 24.48 & 26.01 & 27.31 & \underline{27.57} & \textbf{27.71} \\
& SSIM$\uparrow$ & 0.874 & 0.904 & 0.916 & 0.911 & 0.917 & \underline{0.931} & 0.913 & \textbf{0.934} \\
& LPIPS$\downarrow$ & 0.172 & 0.141 & 0.122 & 0.128 & 0.122 & \underline{0.110} & 0.119 & \textbf{0.092} \\
& MANIQA$\uparrow$ & 0.497 & 0.479 & 0.505 & 0.511 & \underline{0.512} & 0.487 & 0.470 & \textbf{0.513} \\
& QUALICLIP$\uparrow$ & 0.643 & 0.632 & 0.657 & \underline{0.669} & 0.662 & 0.658 & 0.600 & \textbf{0.679} \\
\midrule
\multirow{5}{*}{\begin{tabular}{c}Real20\\(20)\end{tabular}}
& PSNR$\uparrow$ & 21.54 & 22.59 & 23.66 & 21.23 & 23.04 & \textbf{25.23} & \underline{24.30} & 23.74 \\
& SSIM$\uparrow$ & 0.778 & 0.795 & 0.804 & 0.772 & 0.815 & \textbf{0.840} & 0.797 & \underline{0.824} \\
& LPIPS$\downarrow$ & 0.207 & 0.198 & 0.172 & 0.210 & 0.178 & \underline{0.161} & 0.198 & \textbf{0.160} \\
& MANIQA$\uparrow$ & 0.461 & 0.450 & 0.475 & 0.464 & \underline{0.480} & 0.440 & 0.443 & \textbf{0.483} \\
& QUALICLIP$\uparrow$ & 0.677 & 0.665 & 0.694 & \underline{0.719} & 0.684 & 0.672 & 0.635 & \textbf{0.729} \\
\midrule
\multirow{5}{*}{\begin{tabular}{c}DRR-C\\(200)\end{tabular}}
& PSNR$\uparrow$ & 20.44 & 21.14 & 21.41 & 20.38 & 21.65 & \underline{23.83} & 22.64 & \textbf{25.02} \\
& SSIM$\uparrow$ & 0.725 & 0.731 & 0.718 & 0.736 & 0.730 & \underline{0.840} & 0.709 & \textbf{0.919} \\
& LPIPS$\downarrow$ & 0.277 & 0.274 & 0.280 & 0.213 & 0.272 & \underline{0.170} & 0.292 & \textbf{0.117} \\
& MANIQA$\uparrow$ & 0.458 & 0.446 & 0.470 & \underline{0.480} & 0.473 & 0.451 & 0.466 & \textbf{0.490} \\
& QUALICLIP$\uparrow$ & 0.739 & 0.726 & 0.732 & 0.744 & 0.740 & \underline{0.746} & 0.726 & \textbf{0.782} \\
\midrule
\multirow{5}{*}{\begin{tabular}{c}DRR-S\\(200)\end{tabular}}
& PSNR$\uparrow$ & 19.79 & 22.17 & 22.70 & 20.71 & 23.17 & \underline{27.21} & 25.79 & \textbf{28.35} \\
& SSIM$\uparrow$ & 0.775 & 0.794 & 0.802 & 0.802 & 0.808 & \underline{0.901} & 0.811 & \textbf{0.942} \\
& LPIPS$\downarrow$ & 0.246 & 0.228 & 0.224 & 0.214 & 0.217 & \underline{0.119} & 0.215 & \textbf{0.080} \\
& MANIQA$\uparrow$ & 0.520 & 0.484 & 0.508 & \underline{0.524} & 0.514 & 0.497 & 0.496 & \textbf{0.553} \\
& QUALICLIP$\uparrow$ & 0.723 & 0.706 & 0.723 & 0.726 & 0.726 & \underline{0.728} & 0.711 & \textbf{0.731} \\
\midrule
\multirow{5}{*}{\begin{tabular}{c}OpenRR-1k\\(100)\end{tabular}}
& PSNR$\uparrow$ & 22.15 & 23.83 & 24.48 & 20.65 & 23.73 & 25.27 & \underline{25.58} & \textbf{28.27} \\
& SSIM$\uparrow$ & 0.831 & 0.833 & 0.830 & 0.822 & 0.833 & \underline{0.854} & 0.792 & \textbf{0.950} \\
& LPIPS$\downarrow$ & 0.197 & 0.195 & 0.200 & 0.203 & 0.188 & \underline{0.183} & 0.218 & \textbf{0.062} \\
& MANIQA$\uparrow$ & 0.456 & 0.432 & 0.456 & 0.449 & \underline{0.458} & 0.438 & 0.453 & \textbf{0.470} \\
& QUALICLIP$\uparrow$ & \underline{0.724} & 0.695 & 0.715 & 0.715 & 0.723 & 0.709 & 0.714 & \textbf{0.742} \\
\midrule
\multirow{5}{*}{\begin{tabular}{c}Average\\(1040)\end{tabular}}
& PSNR$\uparrow$ & 21.14 & 23.43 & 23.99 & 21.42 & 23.92 & \underline{25.98} & 25.48 & \textbf{26.59} \\
& SSIM$\uparrow$ & 0.791 & 0.815 & 0.817 & 0.809 & 0.822 & \underline{0.869} & 0.810 & \textbf{0.906} \\
& LPIPS$\downarrow$ & 0.221 & 0.202 & 0.198 & 0.201 & 0.192 & \underline{0.159} & 0.208 & \textbf{0.106} \\
& MANIQA$\uparrow$ & 0.479 & 0.458 & 0.479 & 0.485 & \underline{0.486} & 0.460 & 0.465 & \textbf{0.504} \\
& QUALICLIP$\uparrow$ & 0.704 & 0.689 & 0.706 & \underline{0.714} & 0.707 & 0.696 & 0.674 & \textbf{0.731} \\
\bottomrule
\end{tabular}
\end{table*}

\begin{figure*}[tb]
\centering
\setlength{\tabcolsep}{1pt}
\renewcommand{\arraystretch}{0.3}
\begin{tabular}{ccccc}
\scriptsize Input & \scriptsize GT & \scriptsize Ours & \scriptsize WindowSeat & \scriptsize DAI \\
\includegraphics[width=\imgwidth]{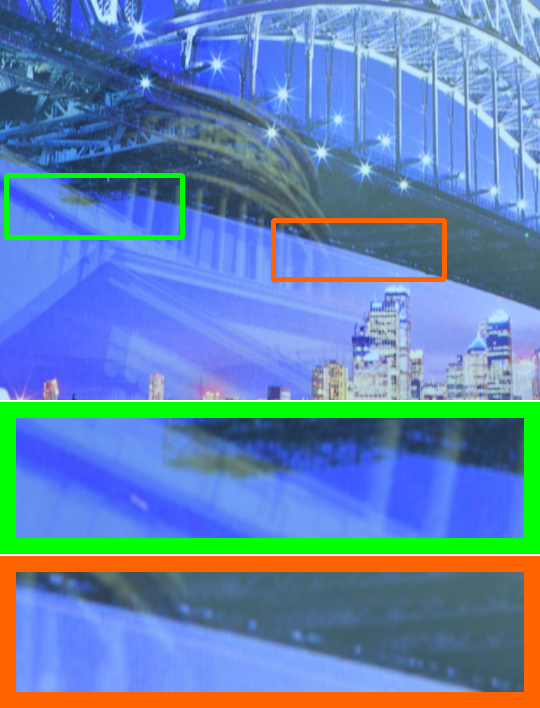} &
\includegraphics[width=\imgwidth]{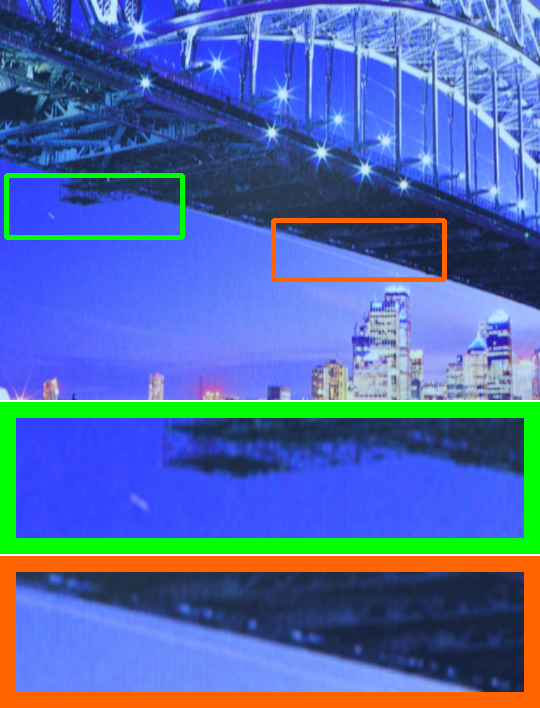} &
\includegraphics[width=\imgwidth]{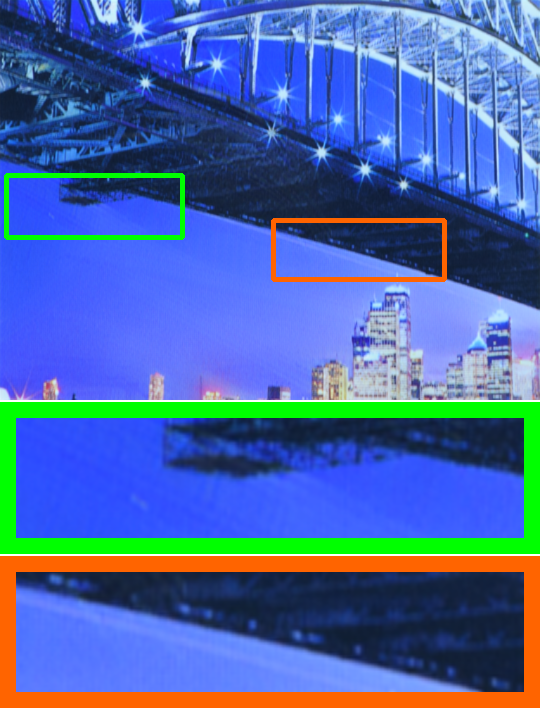} &
\includegraphics[width=\imgwidth]{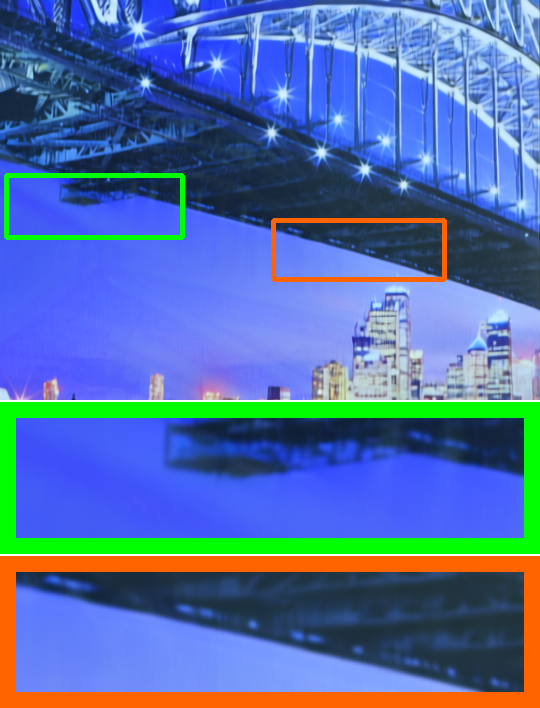} &
\includegraphics[width=\imgwidth]{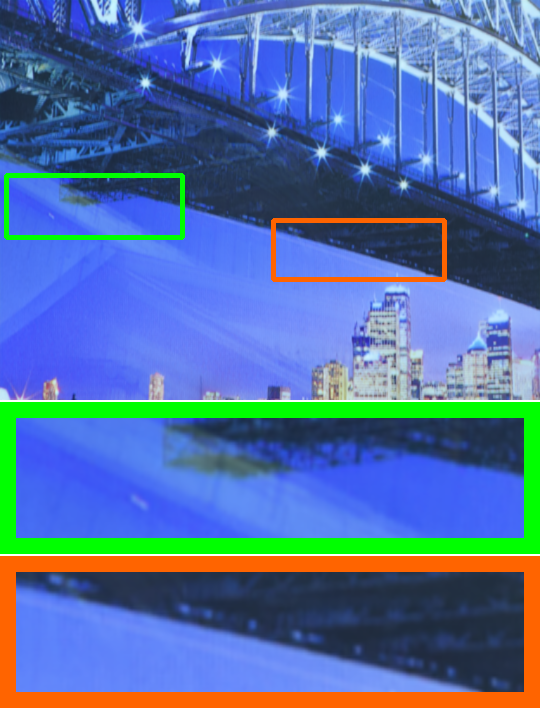} \\
\scriptsize DSIT & \scriptsize RDNet & \scriptsize RobustSIRR & \scriptsize RRW & \scriptsize DSRNet \\
\includegraphics[width=\imgwidth]{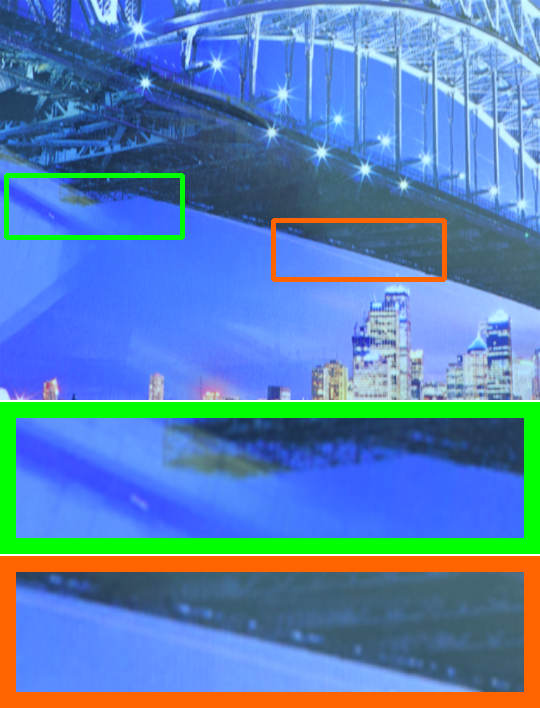} &
\includegraphics[width=\imgwidth]{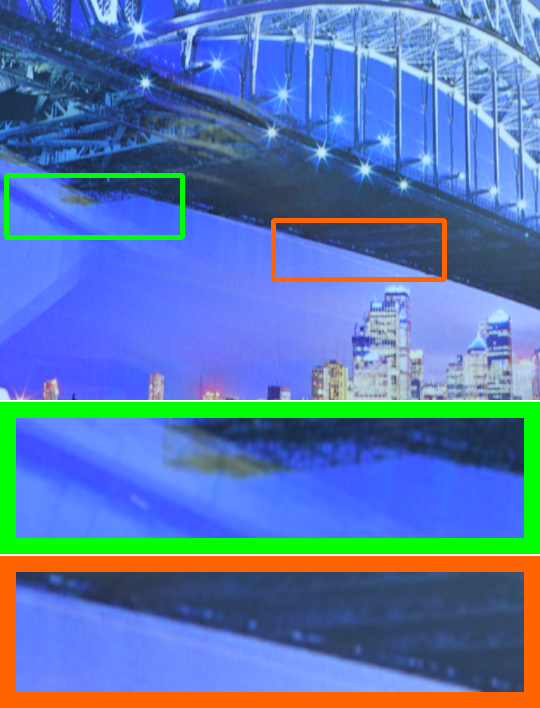} &
\includegraphics[width=\imgwidth]{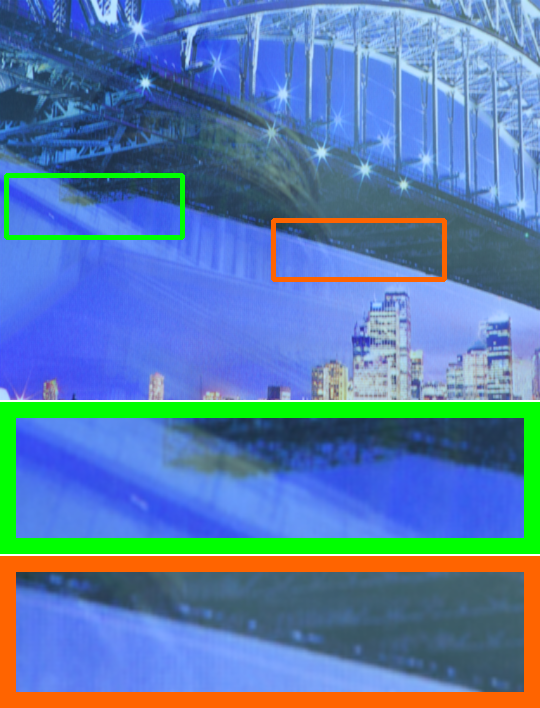} &
\includegraphics[width=\imgwidth]{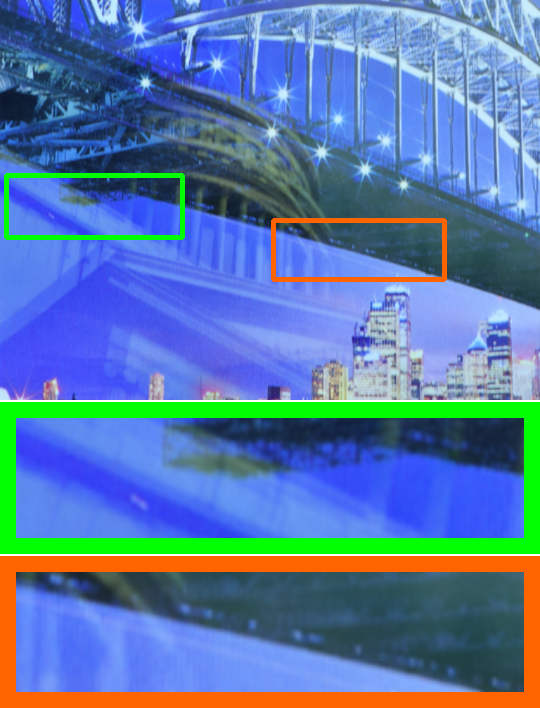} &
\includegraphics[width=\imgwidth]{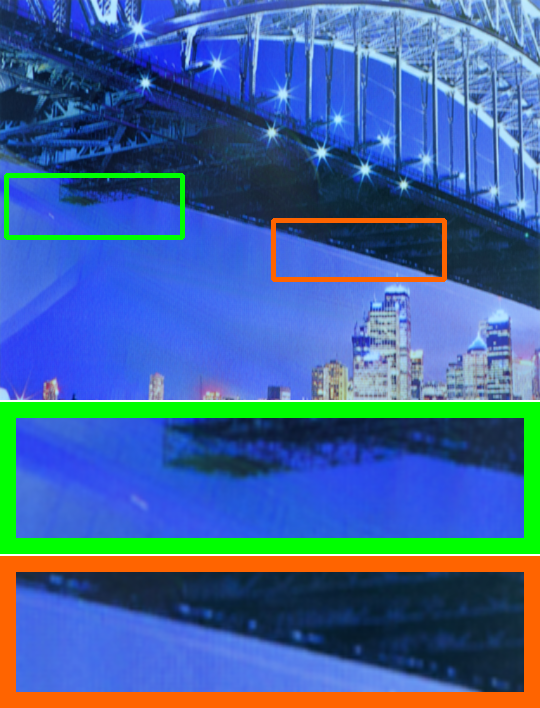} \\
% \noalign{\vskip 3pt}
\scriptsize Input & \scriptsize GT & \scriptsize Ours & \scriptsize WindowSeat & \scriptsize DAI \\
\includegraphics[width=\imgwidth]{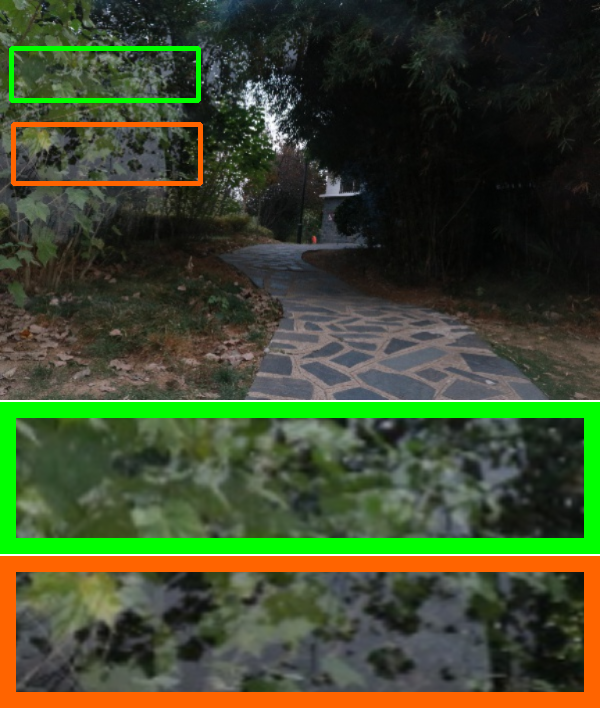} &
\includegraphics[width=\imgwidth]{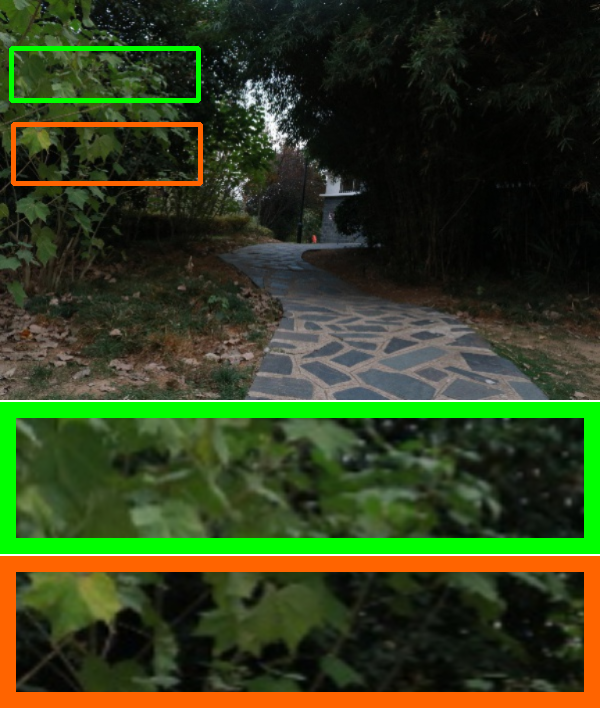} &
\includegraphics[width=\imgwidth]{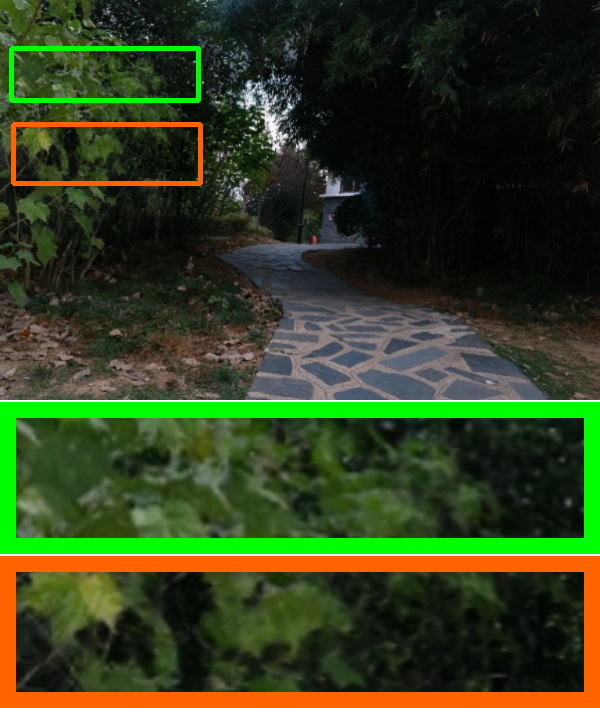} &
\includegraphics[width=\imgwidth]{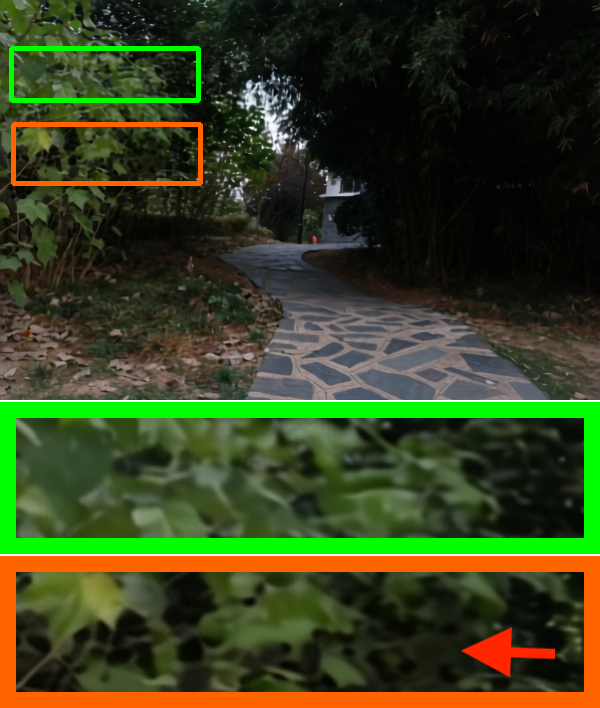} &
\includegraphics[width=\imgwidth]{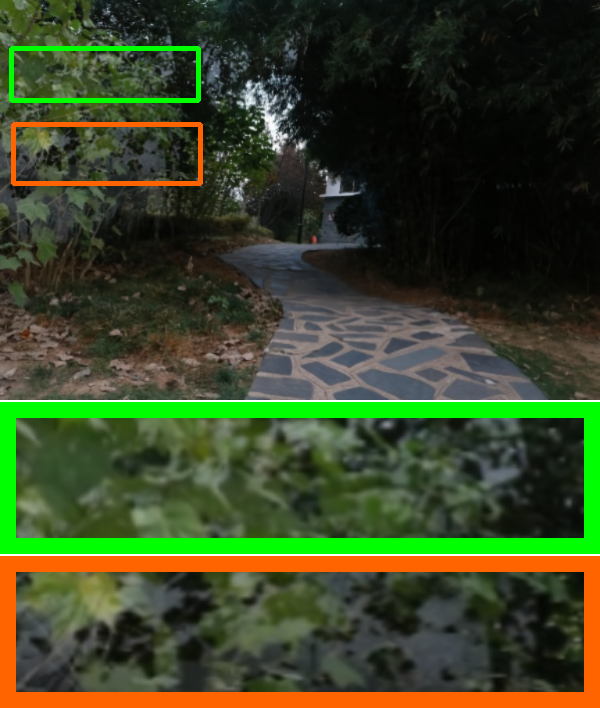} \\
\scriptsize DSIT & \scriptsize RDNet & \scriptsize RobustSIRR & \scriptsize RRW & \scriptsize DSRNet \\
\includegraphics[width=\imgwidth]{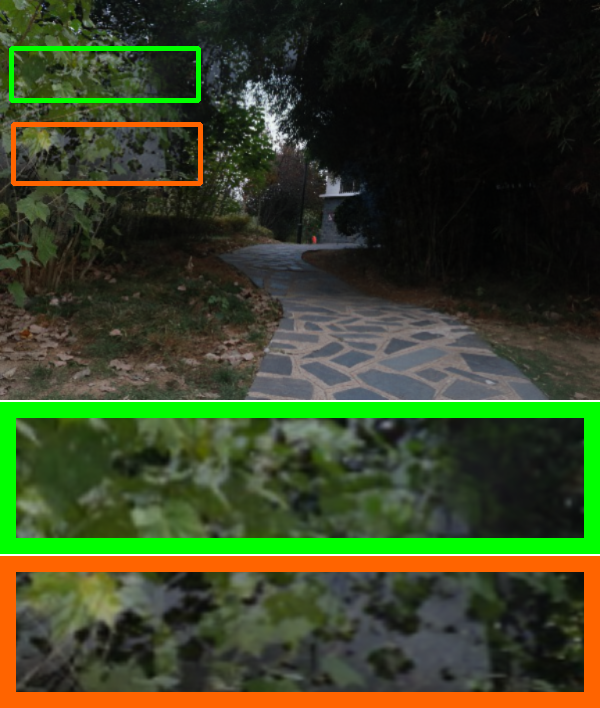} &
\includegraphics[width=\imgwidth]{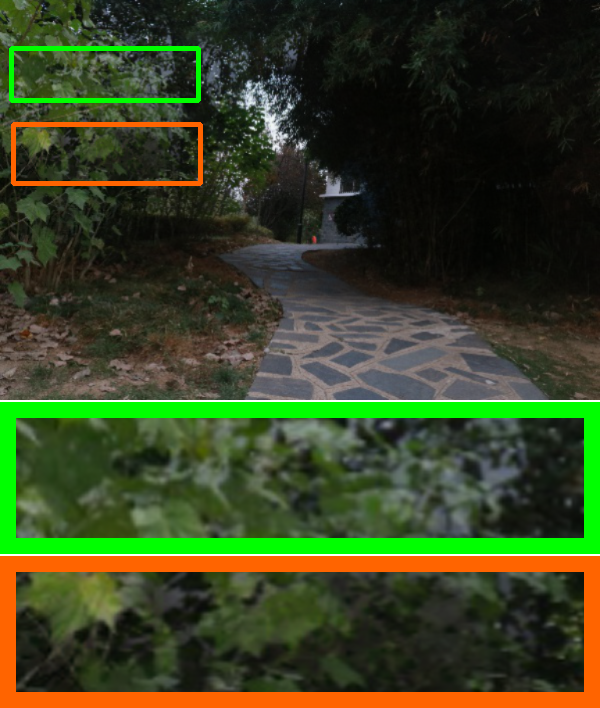} &
\includegraphics[width=\imgwidth]{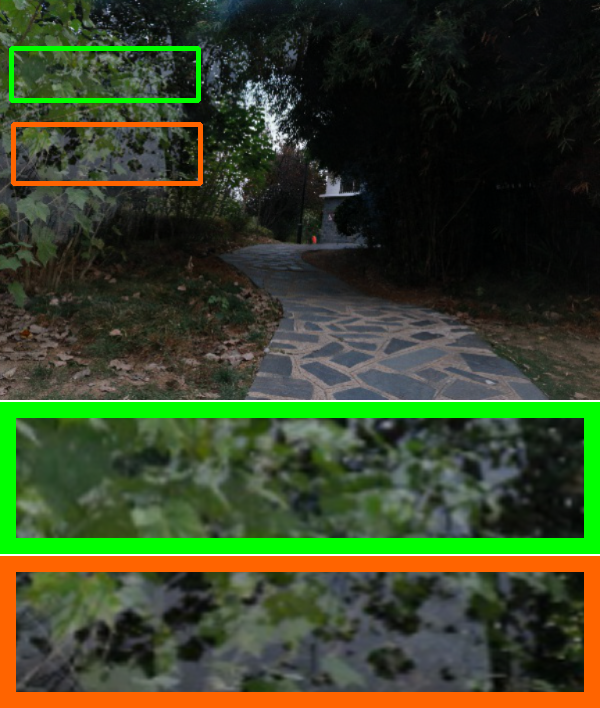} &
\includegraphics[width=\imgwidth]{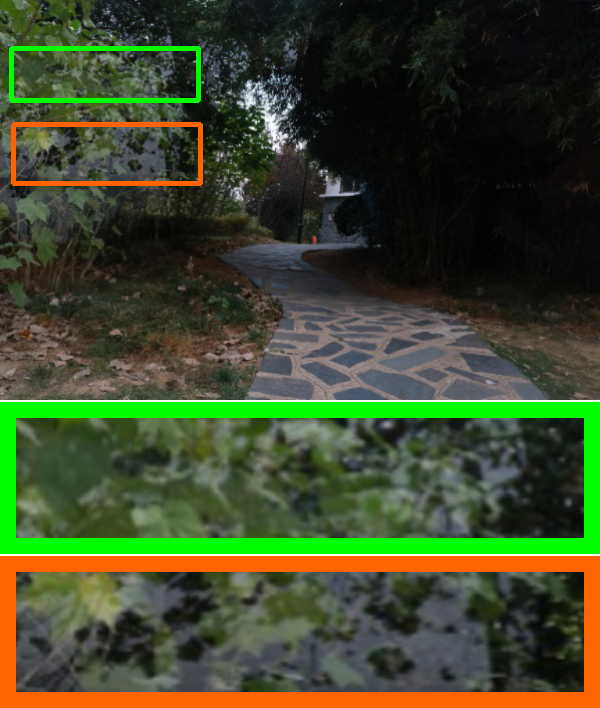} &
\includegraphics[width=\imgwidth]{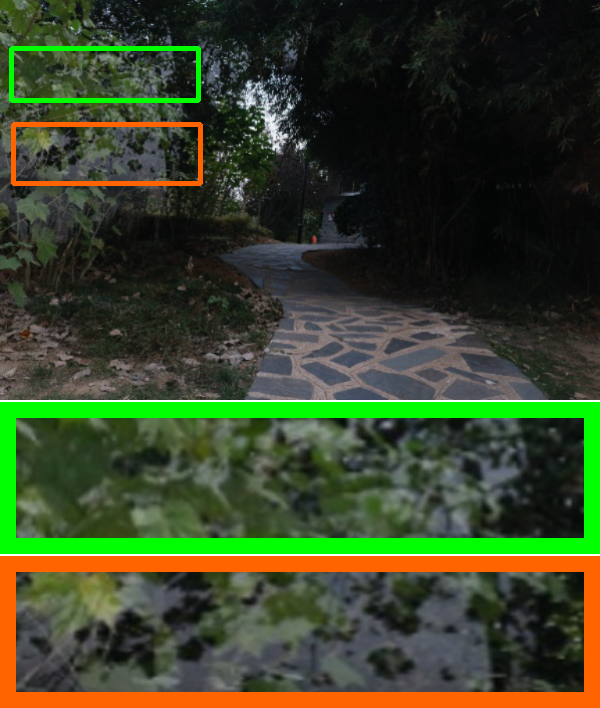} \\
\end{tabular}
\caption{Visual comparison against all methods on SIR$^2$~\cite{SIR2} (top two rows) and Nature~\cite{li2020single} (bottom two rows). Our method yields cleaner separation and recovers structural detail and texture more faithfully. Zoom in for detail.}
\label{fig:comparison_postcard}
\end{figure*}

We implement CLEAR in PyTorch~\cite{pytorch} with a two-stage training strategy on 84{,}425 pairs matching DAI's composition~\cite{hu2025dereflection} for a controlled comparison: DRR~\cite{hu2025dereflection} (23{,}303), synthetic COCO~\cite{coco}/PASCAL VOC~\cite{pascal} samples (60{,}833) generated following Hu~\etal~\cite{hu2023single}, and Nature~\cite{li2020single} (200) plus Zhang~\etal~\cite{zhang2018single} (89) real training splits. Both U-Net branches and the VAE start from pre-trained SDv2 checkpoints~\cite{rombach2022high}; HFAM is randomly initialized. The VAE encoder stays frozen, and Stage 1 remains frozen during Stage 2 training; only the decoder $D$ and HFAM are fine-tuned.

\vspace{-1mm}
\section{Experiments}
\vspace{-1mm}
\label{sec:exp}

\subsection{Training and Inference}
The two stages are trained sequentially on a single NVIDIA A6000 with AdamW~\cite{adamw} (lr $1\times10^{-5}$, batch size 2), at $512^2$ (Stage 1) and $768^2$ (Stage 2) with standard augmentation; each stage runs $\sim$100K iterations. We set $\lambda_T=\lambda_R=\lambda_{rec}=\lambda_{exc}=0.25$ and $\lambda_1=\lambda_2=0.5$. At inference the pipeline is feed-forward and ground-truth-free. At 768$^2$, CLEAR performs four encoder forwards, two U-Net-branch forwards, and two decoder forwards (the cross-latent decoder for $T$; the original VAE decoder with $R$/$\Phi$ batched): $\approx$16.0T MACs (31.9T FLOPs), 0.556\,s on one A6000.

\vspace{-1mm}
\subsection{Comparison with State of the Art}
We evaluate on six benchmarks: \textbf{Nature}~\cite{li2020single}, \textbf{SIR$^2$}~\cite{SIR2}, \textbf{Real20}~\cite{zhang2018single}, \textbf{DRR-C}, \textbf{DRR-S}~\cite{hu2025dereflection}, and \textbf{OpenRR-1k}~\cite{openrr1k}. We compare CLEAR against RobustSIRR~\cite{song2023robust}, RRW~\cite{zhu2024revisiting}, DSIT~\cite{hu2024single}, DSRNet~\cite{hu2023single}, RDNet~\cite{zhao2025reversible}, DAI~\cite{hu2025dereflection}, and WindowSeat~\cite{windowseat2025}. Following the adaptation protocol of DAI~\cite{hu2025dereflection}, \emph{every} method is fine-tuned on our identical training data and evaluated with the same testing scripts, so the comparison isolates architectural and methodological differences rather than training-data advantages. L-DiffER~\cite{hong2024ldiffer} and DPIT~\cite{huang2025single} have no public code at the time of writing and cannot be reproduced. GenSIRR~\cite{gensirr} is concurrent; we run its released checkpoint without refitting, as it likewise does not refit prior methods, and report it separately in \cref{tab:capacity}. We report PSNR and SSIM for fidelity and LPIPS~\cite{lpips}, MANIQA~\cite{maniqa}, and QUALICLIP~\cite{qualiclip} for perceptual quality.

\noindent\textbf{Quantitative results.}
\cref{tab:quantitative_comparison} shows that CLEAR has the best average across all five metrics and the lowest LPIPS on every dataset, with the largest SSIM and perceptual gains on the harder DRR-C, DRR-S, and OpenRR-1k benchmarks. Some competitors reach higher PSNR on individual datasets (\eg DAI on Real20), but PSNR alone does not capture perceived quality for this ill-posed task; the no-reference MANIQA and QUALICLIP scores, closest to ground truth among all methods (supplementary material), corroborate LPIPS.

\subsection{Qualitative Results}
\vspace{-2mm}
\cref{fig:comparison_postcard} visually compares the outputs. CNN-based methods~\cite{song2023robust, hu2023single, zhu2024revisiting, hu2024single, zhao2025reversible} and DAI~\cite{hu2025dereflection} often leave high-frequency artifacts or over-smoothed textures when reflections are strongly entangled with the scene, while WindowSeat~\cite{windowseat2025} can produce plausible but incorrect content. Our deterministic two-stage method removes reflections more cleanly and keeps structure and color closer to the ground truth. In the supplementary material (Reflection Layer Disentanglement), we compare predicted reflection layers: CLEAR isolates reflection texture with less transmission leakage than the pixel-space dual-stream methods (DSRNet, DSIT, RDNet), including DSRNet, whose residue model it adapts to the latent space.
% \cref{fig:reflection_main} compares predicted reflection layers: CLEAR isolates reflection texture with less transmission leakage than the pixel-space dual-stream methods, including DSRNet, whose residue model it adapts to the latent space (Additional predicted reflection visualizations are provided in the supplementary document).

\subsection{Analysis and Ablation Studies}
\vspace{-2mm}
\label{sec:ablation}
We ablate the pipeline on \textbf{Nature}, \textbf{Real20}, and \textbf{SIR}$^2$ (\cref{tab:ablation}) and check whether the gains stem from scale or from the method. The supplement reports Stage 1 branch/init, reconstruction, exclusive, and cross-attention ablations, plus no-reference IQA calibration.

\begin{table}[t]
\centering
\caption{Ablation of \textbf{Stage 2} refinement, the \textbf{Residue} ($\Phi$), and \textbf{HFAM} sub-paths (\textbf{HF} high-frequency, \textbf{Color}) across three benchmarks. Best in bold.}
\label{tab:ablation}
\small
\setlength{\tabcolsep}{4pt}
\renewcommand{\arraystretch}{1.0}
\resizebox{\columnwidth}{!}{%
\begin{tabular}{cc|cc|cccccc}
\toprule
\multirow{2}{*}{\textbf{S2}} & \multirow{2}{*}{$\Phi$} & \multicolumn{2}{c|}{\textbf{HFAM}} & \multicolumn{2}{c}{\textit{Nature}} & \multicolumn{2}{c}{\textit{Real20}} & \multicolumn{2}{c}{\textit{SIR}$^2$} \\
\cmidrule(lr){3-4} \cmidrule(lr){5-6} \cmidrule(lr){7-8} \cmidrule(lr){9-10}
& & HF & Color & PSNR & SSIM & PSNR & SSIM & PSNR & SSIM \\
\midrule
$\times$    & $\times$   & $\times$   & $\times$   & 25.09 & 0.838 & 23.52 & 0.779 & 26.34 & 0.904 \\
\checkmark & $\times$   & $\times$   & $\times$   & 25.32 & 0.847 & 23.52 & 0.819 & 26.41 & 0.932 \\
\checkmark & \checkmark & $\times$   & $\times$   & 26.02 & 0.865 & 23.60 & 0.824 & 26.57 & 0.933 \\
\checkmark & \checkmark & $\times$   & \checkmark & 26.15 & \textbf{0.869} & 23.52 & 0.823 & 26.65 & \textbf{0.934} \\
\checkmark & \checkmark & \checkmark & $\times$   & 26.04 & 0.868 & 23.73 & 0.823 & 26.69 & 0.932 \\
\checkmark & \checkmark & \checkmark & \checkmark & \textbf{26.45} & 0.867 & \textbf{23.74} & \textbf{0.824} & \textbf{27.71} & \textbf{0.934} \\
\bottomrule
\end{tabular}%
}
\end{table}

\noindent\textbf{Stage 2 refinement.}
Stage 1 alone (row 1) separates the layers structurally but leaves blurred textures and residual artifacts. Adding the Stage 2 refinement path raises PSNR across all datasets and confirms that the two-stage design is needed; the qualitative effect is shown in the supplement.

\noindent\textbf{Residue term ($\Phi$).}
Adding the learned residue $\Phi$ (row 3) lets the network account for non-linear effects such as saturation, improving structural fidelity, especially on Real20 and Nature where the linear model $\mathbf{I}=\mathbf{T}+\mathbf{R}$ is often violated. A residual term adds degrees of freedom, but following~\cite{hu2023single} we find it easier to estimate than a per-pixel alpha map and it empirically beats the strictly linear model.

\noindent\textbf{HFAM and its sub-paths.}
The full HFAM (last row) is best on every benchmark while matching or staying close to the best SSIM. The two paths are complementary: from the Stage-2 + $\Phi$ baseline, the color and high-frequency paths each improve PSNR on their own, and combining them is best (\cref{tab:ablation}). This indicates HFAM does more than global color correction: the two paths are the query and value of one attention operator; each is weak alone, and only together do they amplify transmission-colored channels while attenuating reflection-colored ones. The effect is largest on SIR$^2$, where reflections are strong and chromatically distinct (mechanism detailed in the supplementary material). On color fidelity, the full model also reduces the CIEDE2000 color-difference metric $\Delta E_{00}$~\cite{sharma2005ciede2000} by 4.2\% over the Stage-2+$\Phi$ baseline on SIR$^2$ (4.983$\to$4.774), substantiating the color-correction claim with a perceptual color-difference metric. On the 454 SIR$^2$ images with physical reflection GT, the full model also cuts the reflection leakage metric \texttt{leakage\_corr} from 0.2657 to 0.2243 (-15.6\%; paired bootstrap,10k resamples, 95\% CI $[-0.052, -0.031]$; more details in supplement).

\noindent\textbf{Capacity vs.\ methodology.}
We show that the gains come from the method, not the capacity, with four controlled comparisons (\cref{tab:capacity}). CLEAR beats DAI~\cite{hu2025dereflection} when both are trained on the \emph{identical} data (\cref{sec:method}), which rules out a data advantage. It still beats DAI after swapping SDv2 for the lightweight BK-SDM-Small~\cite{bksdm} backbone (1.29B total parameters, below DAI's 1.74B), so the gains cannot come from model size. In Stage 2, HFAM (22.44M) beats a generic refiner matched to its parameters and FLOPs (22.35M; 25.80/0.861/0.152). A half-size HFAM (halved HCCM channels; 26.25/0.903/0.108) also beats that refiner, by $+0.45$\,dB PSNR, $+0.042$ SSIM, $-0.044$ LPIPS.
\begin{table}[t]
\centering
\setlength{\textfloatsep}{10pt} % reduce space between table and main text
\caption{Capacity, inference cost, and quality (averaged over all benchmarks; time on one A6000). $^{\ddagger}$A generic Stage-2 refiner matched to HFAM's parameters (22.35M) and FLOPs, and a half-size HFAM (halved HCCM channels), trained under identical schedules.}
\label{tab:capacity}
\small
\setlength{\tabcolsep}{3pt}
\renewcommand{\arraystretch}{1.0}
\resizebox{\columnwidth}{!}{%
\begin{tabular}{lcccccc}
\toprule
Method & Total & Train. & Time\,(s) & PSNR$\uparrow$ & SSIM$\uparrow$ & LPIPS$\downarrow$ \\
\midrule
DAI~\cite{hu2025dereflection} & 1.74B & 0.95B & 0.40 & 25.98 & 0.869 & 0.159 \\
GenSIRR~\cite{gensirr}$^{\dagger}$ & 12.11B & 12.02B & 53.8 & 26.47 & 0.879 & 0.130 \\
\midrule
Ours (BK-SDM-Small) & 1.29B & 1.08B & 0.42 & 26.24 & \textbf{0.908} & 0.117 \\
Ours (SDv2) & 2.27B & 1.84B & 0.56 & \textbf{26.59} & 0.906 & \textbf{0.106} \\
Ours (generic S2 refiner)$^{\ddagger}$ & 2.27B & 1.84B & 0.57 & 25.80 & 0.861 & 0.152 \\
Ours (half-size HFAM)$^{\ddagger}$ & 2.27B & 1.82B & 0.54 & 26.25 & 0.903 & 0.108 \\
\bottomrule
\end{tabular}%
}
\end{table}

\noindent\textbf{How detail enters Stage 2.}
We compare alternative ways of injecting high-frequency detail into the decoder from the \emph{same} Stage 1 output (\cref{tab:injection}). Copying the input skips directly recovers sharpness over the no-refinement floor (26.21 vs.\ 25.96 PSNR) but reintroduces reflection structure, so it trails HFAM on every metric. Two standard post-hoc color fixes, a low-frequency wavelet fix and an AdaIN-style per-channel mean/std fix~\cite{colorfix}, match only global statistics and fall \emph{below} the no-refinement baseline. HFAM is the best on all three metrics, showing that Stage 2 does more than skip copying or global color correction.

\begin{table}[t]
\centering
\caption{Stage 2 detail-injection alternatives applied to the same Stage 1 output (averaged over six benchmarks). }
\label{tab:injection}
\small
\setlength{\tabcolsep}{4pt}
\begin{tabular}{lccc}
\toprule
Method & PSNR$\uparrow$ & SSIM$\uparrow$ & LPIPS$\downarrow$ \\
\midrule
Stage 1 only (no Stage 2)   & 25.96 & 0.891 & 0.115 \\
Stage 1 + copy input skips  & 26.21 & 0.896 & 0.112 \\
Stage 1 + Wavelet fix & 22.54 & 0.825 & 0.160 \\
Stage 1 + AdaIN fix   & 23.42 & 0.825 & 0.158 \\
Stage 1 + HFAM (ours) & \textbf{26.59} & \textbf{0.906} & \textbf{0.106} \\
\bottomrule
\end{tabular}
\end{table}

\vspace{-1mm}
\section{Conclusion}
\label{sec:conclusion}
\vspace{-2mm}
We presented CLEAR, a deterministic, single-timestep, non-iterative latent framework for single image reflection removal that structures the first stage with the \emph{transmission, reflection, and residue} formation model and aligns input-derived skip detail to the clean transmission with HFAM's color-gated attention, restoring detail while reducing reflection leakage. Parameter-matched and matched-data studies attribute the gains to the method, not scale; six benchmarks show improved perceptual quality and structural fidelity over prior single-timestep baselines.

% \noindent\textbf{Scope and limitations.}
% CLEAR targets the single-image setting and does not use multi-image, flash/no-flash, or polarization cues, so it relies entirely on learned priors. Like any learned solver for this ill-posed task, it can hallucinate where the input is genuinely ambiguous, most visibly when the reflection is nearly opaque or the transmission is severely attenuated (failure cases in the supplement). Working in a compressed latent space also caps the high-frequency detail recoverable at decode, an upper bound HFAM mitigates but cannot remove; extending the alignment principle to multi-cue settings remains future work.
\noindent\textbf{Scope and limitations.}
CLEAR targets the single-image setting and does not use multi-image, flash/no-flash, or polarization cues, so it relies entirely on learned priors. Like any learned solver for this ill-posed task, it can hallucinate where the input is genuinely ambiguous, most visibly when the reflection is nearly opaque or the transmission is severely attenuated (failure cases in the supplement). Working in a compressed latent space also caps the high-frequency detail recoverable at decode, an upper bound that HFAM mitigates but cannot remove. Future work includes extending the alignment principle to multi-cue settings and reducing computational cost: parameter-efficient fine-tuning (e.g., LoRA~\cite{lora}) could reduce training memory, while distillation to smaller backbones could reduce inference cost. Maintaining CLEAR's perceptual gains while improving pixel fidelity on Nature and Real20 under these changes is left to future work.

\end{document}